\documentclass[format=acmsmall, review=false, screen=true]{acmart}

\usepackage{booktabs} 
\usepackage[ruled]{algorithm2e} 
\usepackage{amsmath}
\usepackage{cases}
\usepackage{bbm}

\SetAlFnt{\small}
\SetAlCapFnt{\small}
\SetAlCapNameFnt{\small}
\SetAlCapHSkip{0pt}
\IncMargin{-\parindent}

\acmJournal{TALLIP}

\setcopyright{acmlicensed}

\acmDOI{0000001.0000001}

\begin{document}
\title{Ancient-Modern Chinese Translation with a New Large Training Dataset}

\author{Dayiheng Liu}
\orcid{0000-0002-8755-8941}
\affiliation{
  \institution{College of Computer Science, State Key Laboratory of Hydraulics and Mountain River Engineering, Sichuan University}
  \city{Chengdu}
  \postcode{610065}
  \country{China}}
\email{losinuris@gmail.com}

\author{Kexin Yang}
\affiliation{
  \institution{College of Computer Science, Sichuan University}
  \city{Chengdu}
  \country{China}}

\author{Qian Qu}
\affiliation{
  \institution{College of Computer Science, Sichuan University}
  \city{Chengdu}
  \country{China}}

\author{Jiancheng Lv}
\authornote{This is the corresponding author}
\affiliation{
  \institution{College of Computer Science, State Key Laboratory of Hydraulics and Mountain River Engineering, Sichuan University}
  \city{Chengdu}
  \country{China}}
\email{lvjiancheng@scu.edu.cn}

\begin{abstract}
Ancient Chinese brings the wisdom and spirit culture of the Chinese nation. Automatic translation from ancient Chinese to modern Chinese helps to inherit and carry forward the quintessence of the ancients. However, the lack of large-scale parallel corpus limits the study of machine translation in Ancient-Modern Chinese. In this paper, we propose an Ancient-Modern Chinese clause alignment approach based on the characteristics of these two languages. This method combines both lexical-based information and statistical-based information, which achieves 94.2 F1-score on our manual annotation Test set. We use this method to create a new large-scale Ancient-Modern Chinese parallel corpus which contains 1.24M bilingual pairs. To our best knowledge, this is the first large high-quality Ancient-Modern Chinese dataset. Furthermore, we analyzed and compared the performance of the SMT and various NMT models on this dataset and provided a strong baseline for this task.
\end{abstract}

%
%

%
%


\maketitle

\renewcommand\shortauthors{Liu, D. et al}

\section{Introduction}
Ancient Chinese\footnote{The concept of \textit{ancient Chinese} in this paper almost refers to the cultural/historical notion of literary Chinese (called WenYanWen in Chinese).} is the writing language in ancient China. It is a treasure of Chinese culture which brings together the wisdom and ideas of the Chinese nation and chronicles the ancient cultural heritage of China. Learning ancient Chinese not only helps people to understand and inherit the wisdom of the ancients, but also promotes people to absorb and develop Chinese culture.

However, it is difficult for modern people to read ancient Chinese. Firstly, compared with modern Chinese, ancient Chinese is more concise and shorter. The grammatical order of modern Chinese is also quite different from that of ancient Chinese. Secondly, most modern Chinese words are double syllables, while the most of the ancient Chinese words are monosyllabic. Thirdly, there is more than one polysemous phenomenon in ancient Chinese. In addition, manual translation has a high cost. Therefore, it is meaningful and useful to study the automatic translation from ancient Chinese to modern Chinese. Through ancient-modern Chinese translation, the wisdom, talent and accumulated experience of the predecessors can be passed on to more people. 

Neural machine translation (NMT) \cite{bahdanau2014neural,wu2016google,li2017modeling,liu2018towards,vaswani2017attention} has achieved remarkable performance on many bilingual translation tasks. It is an end-to-end learning approach for machine translation, with the potential to show great advantages over the statistic machine translation (SMT) systems. However, NMT approach has not been widely applied to the ancient-modern Chinese translation task. One of the main reasons is the limited high-quality parallel data resource.

The most popular method of acquiring translation examples is bilingual text alignment \cite{kaji1992learning}. This kind of method can be classified into two types: lexical-based and statistical-based. The lexical-based approaches \cite{wang2005chinese,kit2004clause} focus on lexical information, which utilize the bilingual dictionary \cite{nasution2018generalized,finch2017inducing} or lexical features. Meanwhile, the statistical-based approaches \cite{brown1991aligning,gale1993program} rely on statistical information, such as sentence length ratio in two languages and align mode probability. 

However, these methods are designed for other bilingual language pairs that are written in different language characters (e.g. English-French, Chinese-Japanese). The ancient-modern Chinese has some characteristics that are quite different from other language pairs. For example, ancient and modern Chinese are both written in Chinese characters, but ancient Chinese is highly concise and its syntactical structure is different from modern Chinese. The traditional methods do not take these characteristics into account. In this paper, we propose an effective ancient-modern Chinese text alignment method at the level of clause\footnote{The clause alignment is more fine-grained than sentence alignment. In the experiment, a sentence was splitted into clauses when we meet comma, semicolon, period or exclamation mark.} based on the characteristics of these two languages. The proposed method combines both lexical-based information and statistical-based information, which achieves 94.2 F1-score on Test set. Recently, a simple longest common subsequence based approach for ancient-modern Chinese sentence alignment is proposed in \cite{zhang2018automatic}. Our experiments showed that our proposed alignment approach performs much better than their method.

We apply the proposed method to create a large translation parallel corpus which contains $\sim$1.24M bilingual sentence pairs. To our best knowledge, this is the first large high-quality ancient-modern Chinese dataset.\footnote{The dataset in \cite{zhang2018automatic} contains only 57391 sentence pairs, the dataset in \cite{lin2007chinese} only involves 205 ancient-modern Chinese paragraph pairs, and the dataset in \cite{liu2012sentence} only involves one history book.} Furthermore, we test SMT models and various NMT models on the created dataset and provide a strong baseline for this task. 

\section{Creating Large Training Dataset}

\subsection{Overview}
There are four steps to build the ancient-modern Chinese translation dataset: (i) The parallel corpus crawling and cleaning. (ii) The paragraph alignment. (iii) The clause alignment based on aligned paragraphs. (iv) Augmenting data by merging aligned adjacent clauses. The most critical step is the third step. 

\subsection{Clause Alignment}
In the clause alignment step, we combine both statistical-based and lexical-based information to measure the score for each possible clause alignment between ancient and modern Chinese strings. The dynamic programming is employed to further find overall optimal alignment paragraph by paragraph. According to the characteristics of the ancient and modern Chinese languages, we consider the following factors to measure the alignment score $d(s, t)$ between a bilingual clause pair:

\noindent \textbf{Lexical Matching}. The lexical matching score is used to calculate the matching coverage of the ancient clause $s$. It contains two parts: exact matching and dictionary matching. An ancient Chinese character usually corresponds to one or more modern Chinese words. In the first part, we carry out Chinese Word segmentation to the modern Chinese clause $t$. Then we match the ancient characters and modern words in the order from left to right.\footnote{When an ancient character appears in a modern word, we define the character to exact match the word.} In further matching, the words that have been matched will be deleted from the original clauses. 

However, some ancient characters do not appear in its corresponding modern Chinese words. An ancient Chinese dictionary is employed to address this issue. We preprocess the ancient Chinese dictionary\footnote{The dictionary we used come from the website http://guwen.xiexingcun.com.} and remove the stop words. In this dictionary matching step, we retrieve the dictionary definition of each unmatched ancient character and use it to match the remaining modern Chinese words. To reduce the impact of universal word matching, we use Inverse Document Frequency (IDF) to weight the matching words. The lexical matching score is calculated as:
\begin{align}
L(s, t) = \frac{1}{|s|}\sum_{c \in s} \mathbbm{1}_t\big(c\big) + \frac{1}{|s|} \sum_{c \in \hat{s}} min \bigg(1, \beta \sum_k {\rm idf}_k \cdot \mathbbm{1}_{\hat{t}}\big(\hat{d}^c_k\big)\bigg). \label{eq:1}
\end{align}
The above equation is used to calculate the matching coverage of the ancient clause $s$. The first term of equation \eqref{eq:1} represents exact matching score. $|s|$ denotes the length of $s$, $c$ denotes each ancient character in $s$, and the indicator function $ \mathbbm{1}_t\big(c\big)$ indicates whether the character $c$ can match the words in the clause $t$. The second term is dictionary matching score. Here $\hat{s}$ and $\hat{t}$ represent the remaining unmatched strings of $s$ and $t$, respectively. $\hat{d}^c_k$ denotes the $k$-th character in the dictionary definition of the $c$ and its IDF score is denoted as ${\rm idf} _k$. The $\beta$ is a predefined parameter which is used to normalize the IDF score. We tuned the value of this parameter on the Dev set. 

\begin{figure*}[h]
   \centering
   \includegraphics[width=5in]{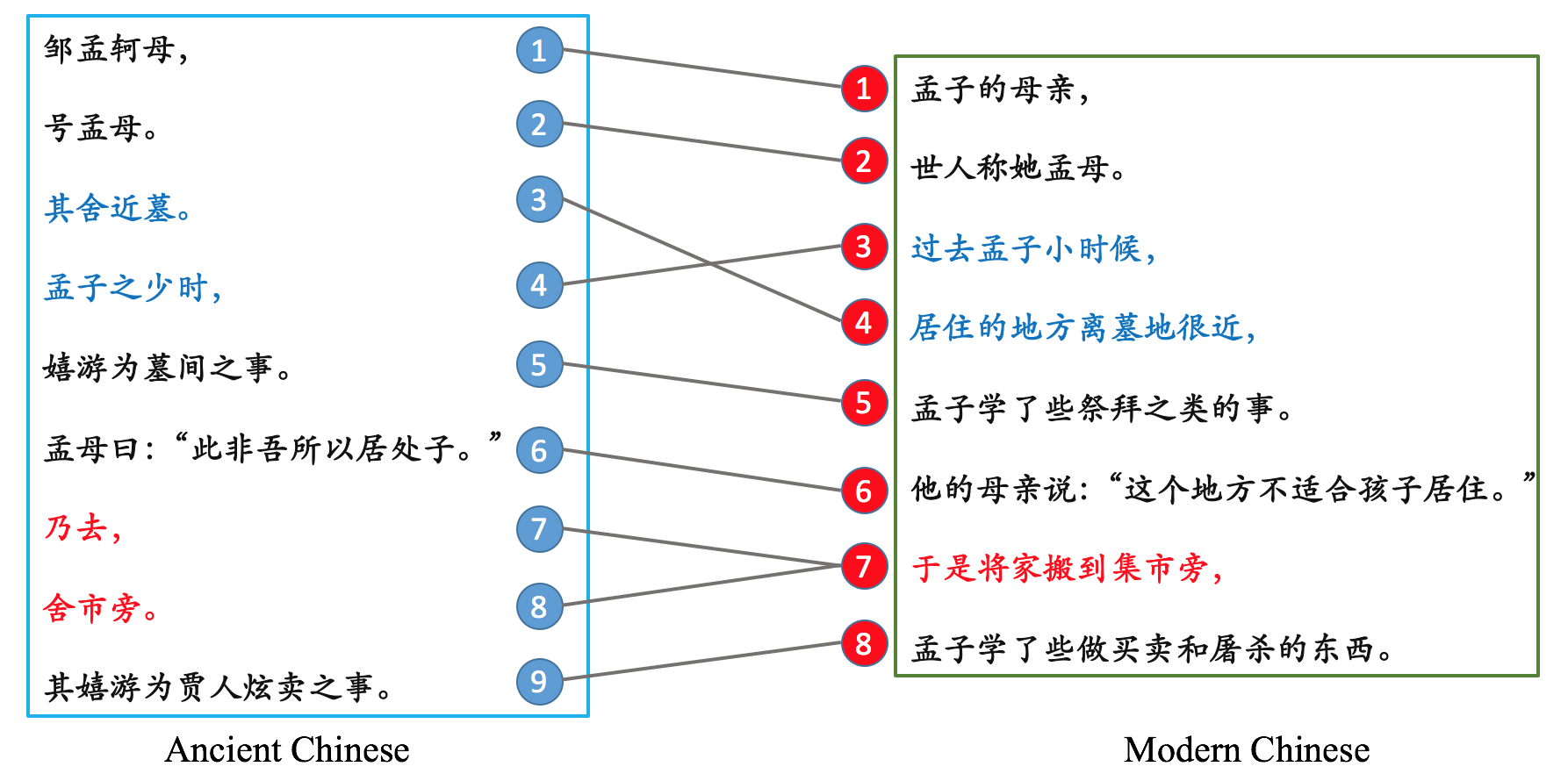}
   \caption{Some examples of different alignment modes. There is a pair of aligned ancient-modern Chinese paragraphs. The ancient Chinese paragraph on the left contains 9 clauses, and the modern Chinese paragraph on the right contains 8 clauses. The lines represent the alignment relation of ancient and modern clauses. The alignment of blue clauses is the 2-2 alignment mode which covers the cases of sentence disordering. And the alignment of red clauses is the 2-1 alignment mode. The rest are in 1-1 alignment mode. [best viewed in color]}
   \label{fig:figure3}
\end{figure*}

\noindent \textbf{Statistical Information}. Similar to \cite{gale1993program} and \cite{wang2005chinese}, the statistical information contains alignment mode and length information. There are many alignment modes between ancient and modern Chinese languages. If one ancient Chinese clause aligns two adjacent modern Chinese clauses, we call this alignment as 1-2 alignment mode. We show some examples of different alignment modes in Figure \ref{fig:figure3}. In this paper, we only consider 1-0, 0-1, 1-1, 1-2, 2-1 and 2-2 alignment modes which account for $99.4\%$ of the Dev set. We estimate the probability Pr$($n-m$)$ of each alignment mode n-m on the Dev set. To utilize length information, we make an investigation on length correlation between these two languages. Based on the assumption of \cite{gale1993program} that each character in one language gives rise to a random number of characters in the other language and those random variables $\delta$ are independent and identically distributed with a normal distribution, we estimate the mean $\mu$ and standard deviation $\sigma$ from the paragraph aligned parallel corpus. Given a clause pair $(s, t)$, the statistical information score can be calculated by:
\begin{equation}
S(s, t) = \varphi \Big(\frac{|s|/|t| - \mu}{\sigma} \Big) \cdot {\rm Pr(n\textrm{-}m)},
\end{equation}
where $\varphi(\cdot)$ denotes the normal distribution probability density function.

\noindent  \textbf{Edit Distance}. Because ancient and modern Chinese are both written in Chinese characters, we also consider using the edit distance. It is a way of quantifying the dissimilarity between two strings by counting the minimum number of operations (insertion, deletion, and substitution) required to transform one string into the other. Here we define the edit distance score as:
\begin{equation}
E(s, t) =  1 - \frac{ {\rm EditDis}(s, t)}{max(|s|, |t|)}.
\end{equation}

\noindent \textbf{Dynamic Programming}.
The overall alignment score for each possible clause alignment is as follows:
\begin{equation}
\normalsize
d(s, t) =  L(s, t) + \gamma S(s, t)  + \lambda E(s, t).
\normalsize
\end{equation}
Here $\gamma$ and $\lambda$ are pre-defined interpolation factors. We use dynamic programming to find the overall optimal alignment paragraph by paragraph. Let $D(i,j)$ be total alignment scores of aligning the first to $i$-th ancient Chinese clauses with the first to to $j$-th modern Chinese clauses, and the recurrence then can be described as follows:

\begin{align}
D(i,j) = max \begin{cases}
 & \text D(i-1,j)+d(s_{i},NULL)\\ 
 & \text D(i,j-1)+d(NULL,t_{j})\\ 
 & \text D(i-1,j-1)+d(s_{i},t_{j})\\ 
 & \text D(i-1,j-2)+d(s_{i},t_{j} \oplus t_{j-1})\\ 
 & \text D(i-2,j-1)+d(s_{i} \oplus s_{i-1},t_{j}) \\ 
 & \text D(i-2,j-2)+d(s_{i} \oplus s_{i-1},t_{j} \oplus t_{j-1}) \\
\end{cases}
\end{align}

Where $s_{i} \oplus s_{i-1}$ denotes concatenate clause $s_{i-1} $ to clause $s_{i}$. As we discussed above, here we only consider 1-0, 0-1, 1-1, 1-2, 2-1 and 2-2 alignment modes.

\begin{figure*}[h]
   \centering
   \includegraphics[width=4in]{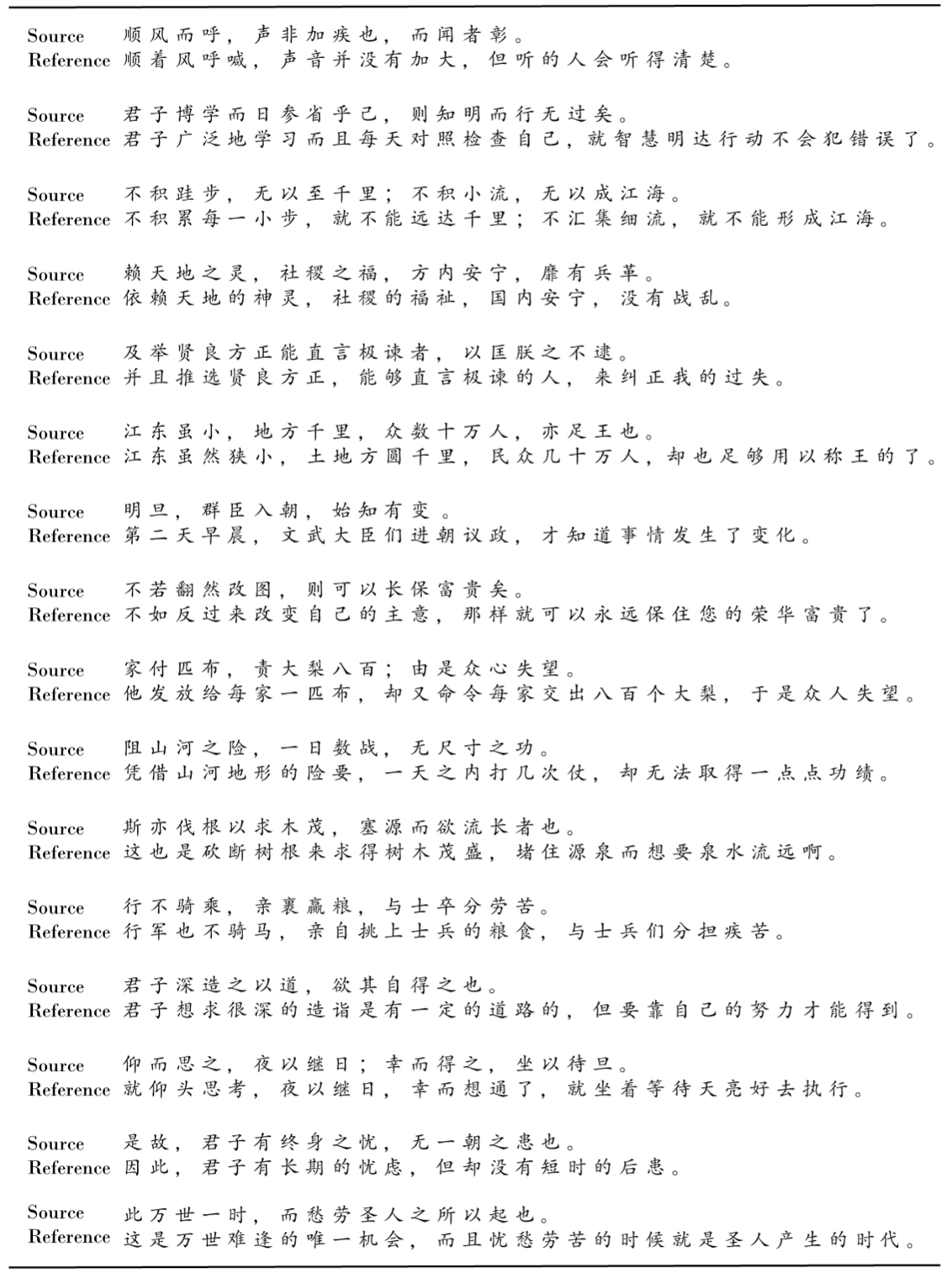}
   \caption{Some data samples of the ancient-modern Chinese parallel corpus. The source language is ancient Chinese and the reference is modern Chinese.}
   \label{fig:figure1}
\end{figure*}

\subsection{Ancient-Modern Chinese Dataset}
\noindent \textbf{Data Collection}. To build the large ancient-modern Chinese dataset, we collected 1.7K bilingual ancient-modern Chinese articles from the internet. More specifically, a large part of the ancient Chinese data we used come from ancient Chinese history records in several dynasties (about 1000BC-200BC) and articles written by celebrities of that era.\footnote{Most of the raw data comes from the website http://www.gushiwen.org and http://wyw.5156edu.com.} They used plain and accurate words to express what happened at that time, and thus ensure the generality of the translated materials. 

\noindent \textbf{Paragraph Alignment}. To further ensure the quality of the new dataset, the work of paragraph alignment is manually completed. After data cleaning and manual paragraph alignment, we obtained 35K aligned bilingual paragraphs.

\noindent \textbf{Clause Alignment}. We applied our clause alignment algorithm on the 35K aligned bilingual paragraphs and obtained 517K aligned bilingual clauses. The reason we use clause alignment algorithm instead of sentence alignment is because we can construct more aligned sentences more flexibly and conveniently. To be specific, we can get multiple additional sentence level bilingual pairs by ``data augmentation''. 

\noindent \textbf{Data Augmentation}. We augmented the data in the following way: Given an aligned clause pair, we merged its adjacent clause pairs as a new sample pair. For example, suppose we have three adjacent clause level bilingual pairs: ($x_1$, $y_1$), ($x_2$, $y_2$), and ($x_3$, $y_3$). We can get some additional sentence level bilingual pairs, such as: ($x_1 \oplus x_2$, $y_1 \oplus y_2$) and ($x_1 \oplus x_2 \oplus x_3$, $y_1 \oplus y_2 \oplus y_3$). Here $x_1$, $x_2$, and $x_3$ are adjacent clauses in the original paragraph, and $a \oplus b$ denotes concatenate clause $b$ to clause $a$. The advantage of using this data augmentation method is that compared with only using ($x_1 \oplus x_2 \oplus x_3$, $y_1 \oplus y_2 \oplus y_3$) as the training data, we can also use ($x_1\oplus x_2$, $y_1 \oplus y_2$) and ($x_2 \oplus x_3$, $y_2 \oplus y_3$) as the training data, which can provide richer supervision information for the model and make the model learn the align information between the source language and the target language better. After the data augmentation, we filtered the sentences which are longer than 50 or contain more than four clause pairs. 

\noindent \textbf{Dataset Creation}. Finally, we split the dataset into three sets: training (Train), development (Dev) and testing (Test). Note that the unaugmented dataset contains 517K aligned bilingual clause pairs from 35K aligned bilingual paragraphs. To keep all the sentences in different sets come from different articles, we split the 35K aligned bilingual paragraphs into Train, Dev and Test sets following these ratios respectively: 80\%, 10\%, 10\%. Before data augmentation, the unaugmented Train set contains $\sim0.46M$ aligned bilingual clause pairs from 28K aligned bilingual paragraphs. Then we augmented the Train, Dev and Test sets respectively. Note that the augmented Train, Dev and Test sets also contain the unaugmented data. The statistical information of the three data sets is shown in Table \ref{data}. We show some examples of data in Figure \ref{fig:figure1}.

\begin{table}[h]
\caption{\label{data} Statistical information of the ancient-modern Chinese parallel corpus.}
\begin{center}
\begin{tabular}{lccc}
 \hline
 Set &  Pairs & Src Token. & Target Token. \\
 \hline 
Train & 1.02M  & 12.56M & 18.35M  \\ 
Dev & 125.71K & 1.51M  & 2.38M \\  
Test &100.63K  & 1.19M  & 1.87M \\  
 \hline
 \end{tabular}
\end{center}
\bigskip\centering
\footnotesize\emph{Src Token:} the number of tokens in source language.
\end{table}

\section{Model}
\subsection{RNN-based NMT model}
We first briefly introduce the RNN based Neural Machine Translation (RNN-based NMT) model. The RNN-based NMT with attention mechanism \cite{bahdanau2014neural} has achieved remarkable performance on many translation tasks. It consists of encoder and decoder part. 

We firstly introduce the encoder part. The input word sequence of source language are individually mapped into a $d$-dimensional vector space $\textbf{X}= [\textbf{x}_1, .., \textbf{x}_T]$. Then a bi-directional RNN \cite{schuster1997bidirectional} with GRU \cite{Cho2014Learning} or LSTM \cite{Hochreiter1997Long} cell converts these vectors into a sequences of hidden states $ [\textbf{h}_1, .., \textbf{h}_T]$. 

For the decoder part, another RNN is used to generate target sequence $[y_1, y_2, ..., y_{T'}]$. The attention mechanism \cite{bahdanau2014neural, luong2015effective} is employed to allow the decoder to refer back to the hidden state sequence and focus on a particular segment. The $i$-th hidden state $\textbf{s}_i$ of decoder part is calculated as:
\begin{equation}
  \textbf{s}_i = \textbf{RNN}(\textbf{g}_i,\textbf{s}_{i-1}).
\end{equation} 
Here \textbf{g}$_i$ is a linear combination of attended context vector \textbf{c}$_i$ and $ \textbf{y}_{i-1}$ is the word embedding of (i-1)-th target word:
\begin{equation}
  \textbf{g}_i = \textbf{W}^y\textbf{y}_{i-1}+ \textbf{W}^c\textbf{c}_i.
\end{equation} 
The attended context vector \textbf{c}$_i$ is computed as a weighted sum of the hidden states of the encoder:
\begin{align}
    & e_{ij} = \textbf{v}_a^T \textbf{tanh}(\textbf{W}^a\textbf{s}_{i-1}+\textbf{U}^a\textbf{h}_j). \\
  & \alpha_{ij} = \frac{exp(e_{ij})}{\sum_k{exp(e_{ik})}}. \\
  & \textbf{c}_i = \sum_j{\alpha_{ij}\textbf{h}_j}.
\end{align} 
The probability distribution vector of the next word $y_i$ is generated according to the following:
\begin{equation}
  \textbf{Pr}(y_i) = \textbf{Softmax}(\textbf{W}^o\textbf{s}_i+b).
\end{equation}
We take this model as the basic RNN-based NMT model in the following experiments.

\subsection{Transformer-NMT}
Recently, the Transformer model \cite{vaswani2017attention} has made remarkable progress in machine translation. This model contains a multi-head self-attention encoder and a multi-head self-attention decoder.

As proposed by \cite{vaswani2017attention}, an attention function maps a query and a set of key-value pairs to an output, where the queries $Q$, keys $K$, and values $V$ are all vectors. The input consists of queries and keys of dimension $d_k$, and values of dimension $d_v$. The attention function is given by:
\begin{equation}
Attention(Q, K, V) = \textbf{Softmax}(\frac{Q K^T}{\sqrt{d_k}})V
\end{equation}

Multi-head attention mechanism projects queries, keys and values to $h$ different representation subspaces and calculates corresponding attention. The attention function outputs are concatenated and projected again before giving the final output. Multi-head attention allows the model to attend to multiple features at different positions.

The encoder is composed of a stack of $N$ identical layers. Each layer has two sub-layers: multi-head self-attention mechanism and position-wise fully connected feed-forward network. Similarly, the decoder is also composed of a stack of $N$ identical layers. In addition to the two sub-layers in each encoder layer, the decoder contains a third sub-layer which performs multi-head attention over the output of the encoder stack (see more details in \cite{vaswani2017attention}).

\section{Experiments}
Our experiments revolve around the following questions: \textbf{Q1:} As we consider three factors for clause alignment, do all these factors help? How does our method compare with previous methods? \textbf{Q2:} How does the NMT and SMT models perform on this new dataset we build?

\subsection{Clause Alignment Results (Q1)}
In order to evaluate our clause alignment algorithm, we manually aligned bilingual clauses from 37 bilingual ancient-modern Chinese articles, and finally got 4K aligned bilingual clauses as the Test set and 2K clauses as the Dev set. 

\begin{table}[h]
\caption{\label{dev} Results of various hyper-parameters on the Dev set. }
\begin{center}
\begin{tabular}{lccccc}
 \hline
 $\beta$ & $\gamma$ & $\lambda$ &  F1 & Precision\\
 \hline 
3 & 0.05 & 0.05  & 94.0 & 94.5\\  
\textbf{5} & 0.05 & 0.05  & \textbf{94.2} & \textbf{94.7}\\ 
10 & 0.05 & 0.05  & 93.8 & 94.3\\
\hline
5 & 0.03 & 0.05  & 93.7 & 94.3\\ 
5 & \textbf{0.05} & 0.05  & \textbf{94.2} & \textbf{94.7}\\ 
5 & 0.10 & 0.05  & 93.6 & 94.2\\ 
\hline
5 & 0.05 & 0.03 & 94.0 & 94.5  \\ 
5 & 0.05 & \textbf{0.05} & \textbf{94.2} & \textbf{94.7}  \\ 
5 & 0.05 & 0.10 & 93.7 & 94.3  \\ 
 \hline
 \end{tabular}
\end{center}
\end{table}

\noindent \textbf{Metrics}. We used F1-score and precision score as the evaluation metrics. Suppose that we get $N$ bilingual clause pairs after running the algorithm on the Test set, and there are $T$ bilingual clause pairs of these $N$ pairs are in the ground truth of the Test set, the precision score is defined as $P=\frac{T}{N}$ (the algorithm gives $N$ outputs, $T$ of which are correct). And suppose that the ground truth of the Test set contains $M$ bilingual clause pairs, the recall score is $R=\frac{T}{M}$ (there are $M$ ground truth samples, $T$ of which are output by the algorithm), then the F1-score is $2\cdot \frac{P \cdot R}{P + R}$.

\noindent \textbf{Baselines}. Since the related work \cite{brown1991aligning,gale1993program} can be seen as the ablation cases of our method (only statistical score $S(s,t)$ with dynamic programming), we compared the full proposed method with its variants on the Test set for ablation study. In addition, we also compared our method with the longest common subsequence (LCS) based approach proposed by \cite{zhang2018automatic}. To the best of our knowledge, \cite{zhang2018automatic} is the latest related work which are designed for Ancient-Modern Chinese alignment. 

\noindent \textbf{Hyper-parameters}. For the proposed method, we estimated $\mu$ and $\sigma$ on all aligned paragraphs. The probability Pr$($n-m$)$ of each alignment mode n-m was estimated on the Dev set. For the hyper-parameters $\beta$, $\gamma$ and $\lambda$, the grid search was applied to tune them on the Dev set. In order to show the effect of hyper-parameters $\beta$, $\gamma$, and $\lambda$, we reported the results of various hyper-parameters on the Dev set in Table \ref{dev}. Based on the results of grid search on the Dev set, we set $\beta=5$, $\gamma=0.05$, and $\lambda=0.05$ in the following experiment. The Jieba Chinese text segmentation\footnote{A Python based Chinese word segmentation module https://github.com/fxsjy/jieba.} is employed for modern Chinese word segmentation. 

\noindent \textbf{Results}. The results on the Test set are shown in Table \ref{align}, the abbreviation \textit{w/o} means removing a particular part from the setting. From the results, we can see that the lexical matching score is the most important among these three factors, and statistical information score is more important than edit distance score. Moreover, the dictionary term in lexical matching score significantly improves the performance. From these results, we obtain the best setting that involves all these three factors. We used this setting for dataset creation. Furthermore, the proposed method performs much better than LCS \cite{zhang2018automatic}.

\begin{table}[h]
\caption{\label{align} Evaluation results on the Test set. }
\begin{center}
\begin{tabular}{lccc}
 \hline
 Setting &  F1 & Precision\\
 \hline 
all & \textbf{94.2} & \textbf{94.8} \\ 
w/o lexical score & 84.3 & 86.5 \\ 
w/o statistical score & 92.8 & 93.9 \\  
w/o edit distance & 93.9 & 94.4  \\ 
w/o dictionary & 93.1 & 93.9 \\
LCS \cite{zhang2018automatic} & 91.3 & 92.2 \\
 \hline
 \end{tabular}
\end{center}
\bigskip\centering
\footnotesize\emph{w/o:} without. \emph{w/o dictionary:} without using dictionary term in lexical matching score.
\end{table}

\subsection{Translation Results (Q2)}
In this experiment, we analyzed and compared the performance of the SMT and various NMT models on our built dataset. To verify the effectiveness of our data augmented method. We trained the NMT and SMT models on both unaugmented dataset (including 0.46M training pairs) and augmented dataset, and test all the models on the same Test set which is augmented.\footnote{For a more realistic application scenario, we want the model to be able to translate at the sentence level, so we test models on the Test set which is augmented.} The models to be tested and their configurations are as follows: 

\noindent \textbf{SMT}. The state-of-art Moses toolkit \cite{koehn2007moses} was used to train SMT model. We used KenLM \cite{Heafield2011KenLM} to train a 5-gram language model, and the GIZA++ toolkit to align the data. 

\noindent \textbf{RNN-based NMT}. The basic RNN-based NMT model is based on \cite{bahdanau2014neural} which is introduced above. Both the encoder and decoder used 2-layer RNN with 1024 LSTM cells\footnote{We also tried a larger number of parameters, but found that the performance did not improve.}, and the encoder is a bi-directional RNN. The batch size, threshold of element-wise gradient clipping and initial learning rate of Adam optimizer \cite{kingma2014adam} were set to 128, 5.0 and 0.001. When trained the model on augmented dataset, we used 4-layer RNN. Several techniques were investigated to train the model, including layer-normalization \cite{ba2016layer}, RNN-dropout \cite{gal2016theoretically}, and learning rate decay \cite{wu2016google}. The hyper-parameters were chosen empirically and adjusted in the Dev set. Furthermore, we tested the basic NMT model with several techniques, such as target language reversal \cite{NIPS2014_5346} (reversing the order of the words in all target sentences, but not source sentences), residual connection \cite{he2016deep} and pre-trained word2vec \cite{mnih2013learning}. For word embedding pre-training, we collected an external ancient corpus which contains $\sim$134M tokens.
 
\begin{table}[t]
\caption{\label{transformer} The training configuration of the Transformer.}
\begin{center}
\begin{tabular}{lcc}
 \hline 
 Setting &  Basic Transformer  & Transformer + Augment\\ 
 \hline
Batch Size  & 32 & 32\\
Inner Hidden Size & 1024 & 2048 \\
Word Embedding Size & 256 & 512\\
Dropout  Rate & 0.1 & 0.1\\
Num Heads $h$ & 8 & 8\\
Num Layers $N$ & 6 & 6\\
$d_k$ & 32 & 64 \\
$d_v$ & 32 & 64 \\
$d_{model}$ & 256 & 512 \\
 \hline
 \end{tabular}
 \end{center}
\end{table}

\noindent \textbf{Transformer-NMT}. We also trained the Transformer model \cite{vaswani2017attention} which is a strong baseline of NMT on both augmented and unaugmented parallel corpus. The training configuration of the Transformer\footnote{The implement of the Transformer model is based on https://github.com/Kyubyong/transformer.} model is shown in Table \ref{transformer}. The hyper-parameters are set based on the settings in the paper \cite{vaswani2017attention} and the sizes of our training sets. 

For the evaluation, we used the average of 1 to 4 gram BLEUs multiplied by a brevity penalty \cite{Papineni2002BLEU} which computed by \textit{multi-bleu.perl} in Moses as metrics. The results are reported in Table \ref{bleu}. For RNN-based NMT, we can see that target language reversal, residual connection, and word2vec can further improve the performance of the basic RNN-based NMT model. However, we find that word2vec and reversal tricks seem no obvious improvement when trained the RNN-based NMT and Transformer models on augmented parallel corpus. For SMT, it performs better than NMT models when they were trained on the unaugmented dataset. Nevertheless, when trained on the augmented dataset, both the RNN-based NMT model and Transformer based NMT model outperform the SMT model. In addition, as with other translation tasks \cite{vaswani2017attention}, the Transformer also performs better than RNN-based NMT. 

Because the Test set contains both augmented and unaugmented data, it is not surprising that the RNN-based NMT model and Transformer based NMT model trained on unaugmented data would perform poorly. In order to further verify the effect of data augmentation, we report the test results of the models on only unaugmented test data (including 48K test pairs) in Table \ref{tab:bleu2}. From the results, it can be seen that the data augmentation can still improve the models.

\begin{table}[h]
\caption{\label{bleu} 1-4 gram BLEUs results on various models.}
\begin{center}
\begin{tabular}{lccccc}
 \hline 
 Model & 1-gram & 2-gram & 3-gram & 4-gram\\
 \hline 
SMT & 50.24 & 38.05 & 29.93 & 24.04 \\ 
\ + Augment  & 51.70 & 39.09  & 30.73 & 24.72 \\
\hline
basic RNN-based NMT  & 23.43 & 17.24  & 13.16 & 10.26\\ 
\ + Reversal   & 29.84 & 21.55 & 16.13 & 12.39 \\
\ + Residual  & 33.66 & 23.52  & 17.25 & 13.03 \\
\ + Word2vec   & 34.29 & 25.11 & 19.02 & 14.74 \\
\ + Augment  & 53.35 & 40.10 & 31.51 & 25.42\\
 \hline
 basic Transformer & 39.51 & 29.14 & 22.42 & 17.72 \\ 
\ + Augment  & \textbf{54.23} & \textbf{41.61} & \textbf{33.22} & \textbf{27.16} \\
\hline
 \end{tabular}
\end{center}
\end{table}

\begin{table}[h]
\caption{\label{tab:bleu2} 1-4 gram BLEUs results of NMT models tested on only unaugmented test data.}
\begin{center}
\begin{tabular}{lccccc}
 \hline 
 Model & 1-gram & 2-gram & 3-gram & 4-gram\\
\hline
basic RNN-based NMT  & 49.67 & 37.00  & 28.68 & 22.64\\ 
\ + Augment  & 49.90 & 38.24 & 30.28 & 24.43\\
 \hline
 basic Transformer & 50.47 & 38.28 & 30.17 & 24.47 \\ 
\ + Augment  & \textbf{51.85} & \textbf{39.60} & \textbf{31.51} & \textbf{25.75} \\
\hline
 \end{tabular}
\end{center}
\end{table}

\begin{figure*}[h]
   \centering
   \includegraphics[width=4in]{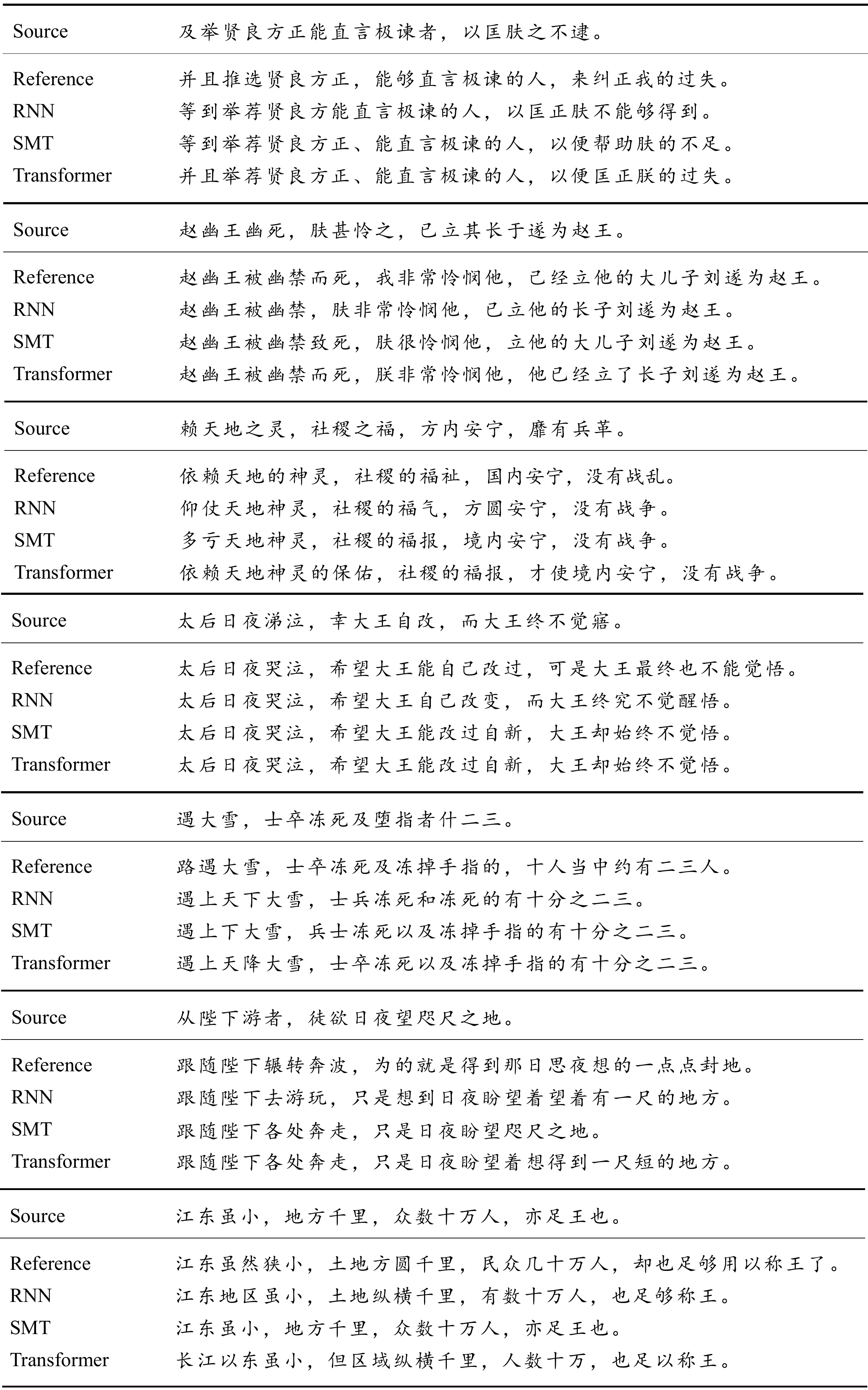}
   \caption{Some generated samples of various models.}
   \label{fig:figure2}
\end{figure*}

\subsection{Analysis}
The generated samples of various models are shown in Figure \ref{fig:figure2}. Besides BLEU scores, we analyze these examples from a human perspective and draw some conclusions. At the same time, we design different metrics and evaluate on the whole Test set to support our conclusions as follows:

On the one hand, we further compare the translation results from the perspective of people. We find that although the original meaning can be basically translated by SMT, its translation results are less smooth when compared with the other two NMT models (RNN-based NMT and Transformer). For example, the translations of SMT are usually lack of auxiliary words, conjunctions and function words, which is not consistent with human translation habits. To further confirm this conclusion, the average length of the translation results of the three models are measured (RNN-based NMT:17.12, SMT:15.50, Transformer:16.78, Reference:16.47). We can see that the average length of the SMT outputs is shortest, and the length gaps between the SMT outputs and the references are largest. Meanwhile, the average length of the sentences translated by Transformer is closest to the average length of references. These results indirectly verify our point of view, and show that the NMT models perform better than SMT in this task.

On the other hand, there still exists some problems to be solved. We observe that translating proper nouns and personal pronouns (such as names, place names and ancient-specific appellations) is very difficult for all of these models. For instance, the ancient Chinese appellation `Zhen' should be translated into `Wo' in modern Chinese. Unfortunately, we calculate the accurate rate of some special words (such as `Zhen',`Chen' and `Gua'), and find that this rate is very low (the accurate rate of translating `Zhen' are: RNN-based NMT:0.14, SMT:0.16, Transformer:0.05). We will focus on this issue in the future.

\section{Conclusion and Future Work}
We propose an effective ancient-modern Chinese clause alignment method which achieves 94.2 F1-score on Test set. Based on it, we build a large scale parallel corpus which contains $\sim$1.24M bilingual sentence pairs. To our best knowledge, this is the first large high-quality ancient-modern Chinese dataset. In addition, we test the performance of the SMT and various NMT models on our built dataset and provide a strong NMT baseline for this task which achieves 27.16 BLEU score (4-gram). We further analyze the performance of the SMT and various NMT models and summarize some specific problems that machine translation models will encounter when translating ancient Chinese.

For the future work, firstly, we are going to expand the dataset using the proposed method continually. Secondly, we will focus on solving the problem of proper noun translation and improve the translation system according to the features of ancient Chinese translation. Finally, we plan to introduce some techniques of statistical translation into neural machine translation to improve the performance.

\begin{acks}
This work is supported by National Natural Science Fund for Distinguished Young Scholar (Grant No. 61625204) and partially supported by the State Key Program of National Science Foundation of China (Grant Nos. 61836006 and 61432014). 
\end{acks}

\bibliographystyle{ACM-Reference-Format}
\bibliography{a2c}


\begin{thebibliography}{28}


\ifx \showCODEN    \undefined \def \showCODEN     #1{\unskip}     \fi
\ifx \showDOI      \undefined \def \showDOI       #1{#1}\fi
\ifx \showISBNx    \undefined \def \showISBNx     #1{\unskip}     \fi
\ifx \showISBNxiii \undefined \def \showISBNxiii  #1{\unskip}     \fi
\ifx \showISSN     \undefined \def \showISSN      #1{\unskip}     \fi
\ifx \showLCCN     \undefined \def \showLCCN      #1{\unskip}     \fi
\ifx \shownote     \undefined \def \shownote      #1{#1}          \fi
\ifx \showarticletitle \undefined \def \showarticletitle #1{#1}   \fi
\ifx \showURL      \undefined \def \showURL       {\relax}        \fi
\providecommand\bibfield[2]{#2}
\providecommand\bibinfo[2]{#2}
\providecommand\natexlab[1]{#1}
\providecommand\showeprint[2][]{arXiv:#2}

\bibitem[\protect\citeauthoryear{Ba, Kiros, and Hinton}{Ba
  et~al\mbox{.}}{2016}]%
        {ba2016layer}
\bibfield{author}{\bibinfo{person}{Jimmy~Lei Ba}, \bibinfo{person}{Jamie~Ryan
  Kiros}, {and} \bibinfo{person}{Geoffrey~E Hinton}.}
  \bibinfo{year}{2016}\natexlab{}.
\newblock \showarticletitle{Layer normalization}.
\newblock \bibinfo{journal}{\emph{arXiv preprint arXiv:1607.06450}}
  (\bibinfo{year}{2016}).
\newblock


\bibitem[\protect\citeauthoryear{Bahdanau, Cho, and Bengio}{Bahdanau
  et~al\mbox{.}}{2014}]%
        {bahdanau2014neural}
\bibfield{author}{\bibinfo{person}{Dzmitry Bahdanau},
  \bibinfo{person}{Kyunghyun Cho}, {and} \bibinfo{person}{Yoshua Bengio}.}
  \bibinfo{year}{2014}\natexlab{}.
\newblock \showarticletitle{Neural machine translation by jointly learning to
  align and translate}.
\newblock \bibinfo{journal}{\emph{arXiv preprint arXiv:1409.0473}}
  (\bibinfo{year}{2014}).
\newblock


\bibitem[\protect\citeauthoryear{Brown, Lai, and Mercer}{Brown
  et~al\mbox{.}}{1991}]%
        {brown1991aligning}
\bibfield{author}{\bibinfo{person}{Peter~F Brown}, \bibinfo{person}{Jennifer~C
  Lai}, {and} \bibinfo{person}{Robert~L Mercer}.}
  \bibinfo{year}{1991}\natexlab{}.
\newblock \showarticletitle{Aligning sentences in parallel corpora}. In
  \bibinfo{booktitle}{\emph{ACL}}.
\newblock


\bibitem[\protect\citeauthoryear{Cho, Van~Merri{\"e}nboer, Gulcehre, Bahdanau,
  Bougares, Schwenk, and Bengio}{Cho et~al\mbox{.}}{2014}]%
        {Cho2014Learning}
\bibfield{author}{\bibinfo{person}{Kyunghyun Cho}, \bibinfo{person}{Bart
  Van~Merri{\"e}nboer}, \bibinfo{person}{Caglar Gulcehre},
  \bibinfo{person}{Dzmitry Bahdanau}, \bibinfo{person}{Fethi Bougares},
  \bibinfo{person}{Holger Schwenk}, {and} \bibinfo{person}{Yoshua Bengio}.}
  \bibinfo{year}{2014}\natexlab{}.
\newblock \showarticletitle{Learning phrase representations using RNN
  encoder-decoder for statistical machine translation}.
\newblock \bibinfo{journal}{\emph{arXiv preprint arXiv:1406.1078}}
  (\bibinfo{year}{2014}).
\newblock


\bibitem[\protect\citeauthoryear{Finch, Harada, Tanaka-Ishii, and Sumita}{Finch
  et~al\mbox{.}}{2017}]%
        {finch2017inducing}
\bibfield{author}{\bibinfo{person}{Andrew Finch}, \bibinfo{person}{Taisuke
  Harada}, \bibinfo{person}{Kumiko Tanaka-Ishii}, {and}
  \bibinfo{person}{Eiichiro Sumita}.} \bibinfo{year}{2017}\natexlab{}.
\newblock \showarticletitle{Inducing a bilingual lexicon from short parallel
  multiword sequences}.
\newblock \bibinfo{journal}{\emph{ACM Transactions on Asian and Low-Resource
  Language Information Processing (TALLIP)}} (\bibinfo{year}{2017}).
\newblock


\bibitem[\protect\citeauthoryear{Gal and Ghahramani}{Gal and
  Ghahramani}{2016}]%
        {gal2016theoretically}
\bibfield{author}{\bibinfo{person}{Yarin Gal} {and} \bibinfo{person}{Zoubin
  Ghahramani}.} \bibinfo{year}{2016}\natexlab{}.
\newblock \showarticletitle{A theoretically grounded application of dropout in
  recurrent neural networks}. In \bibinfo{booktitle}{\emph{NIPS}}.
\newblock


\bibitem[\protect\citeauthoryear{Gale and Church}{Gale and Church}{1993}]%
        {gale1993program}
\bibfield{author}{\bibinfo{person}{William~A Gale} {and}
  \bibinfo{person}{Kenneth~W Church}.} \bibinfo{year}{1993}\natexlab{}.
\newblock \showarticletitle{A program for aligning sentences in bilingual
  corpora}.
\newblock \bibinfo{journal}{\emph{Computational linguistics}}
  (\bibinfo{year}{1993}).
\newblock


\bibitem[\protect\citeauthoryear{He, Zhang, Ren, and Sun}{He
  et~al\mbox{.}}{2016}]%
        {he2016deep}
\bibfield{author}{\bibinfo{person}{Kaiming He}, \bibinfo{person}{Xiangyu
  Zhang}, \bibinfo{person}{Shaoqing Ren}, {and} \bibinfo{person}{Jian Sun}.}
  \bibinfo{year}{2016}\natexlab{}.
\newblock \showarticletitle{Deep residual learning for image recognition}. In
  \bibinfo{booktitle}{\emph{CVPR}}.
\newblock


\bibitem[\protect\citeauthoryear{Heafield}{Heafield}{2011}]%
        {Heafield2011KenLM}
\bibfield{author}{\bibinfo{person}{Kenneth Heafield}.}
  \bibinfo{year}{2011}\natexlab{}.
\newblock \showarticletitle{KenLM: Faster and smaller language model queries}.
  In \bibinfo{booktitle}{\emph{Proceedings of the Sixth Workshop on Statistical
  Machine Translation}}.
\newblock


\bibitem[\protect\citeauthoryear{Hochreiter and Schmidhuber}{Hochreiter and
  Schmidhuber}{1997}]%
        {Hochreiter1997Long}
\bibfield{author}{\bibinfo{person}{Sepp Hochreiter} {and}
  \bibinfo{person}{J{\"u}rgen Schmidhuber}.} \bibinfo{year}{1997}\natexlab{}.
\newblock \showarticletitle{Long short-term memory}.
\newblock \bibinfo{journal}{\emph{Neural computation}} (\bibinfo{year}{1997}).
\newblock


\bibitem[\protect\citeauthoryear{Kaji, Kida, and Morimoto}{Kaji
  et~al\mbox{.}}{1992}]%
        {kaji1992learning}
\bibfield{author}{\bibinfo{person}{Hiroyuki Kaji}, \bibinfo{person}{Yuuko
  Kida}, {and} \bibinfo{person}{Yasutsugu Morimoto}.}
  \bibinfo{year}{1992}\natexlab{}.
\newblock \showarticletitle{Learning translation templates from bilingual
  text}. In \bibinfo{booktitle}{\emph{Computational linguistics}}.
\newblock


\bibitem[\protect\citeauthoryear{Kingma and Ba}{Kingma and Ba}{2014}]%
        {kingma2014adam}
\bibfield{author}{\bibinfo{person}{Diederik~P Kingma} {and}
  \bibinfo{person}{Jimmy Ba}.} \bibinfo{year}{2014}\natexlab{}.
\newblock \showarticletitle{Adam: A method for stochastic optimization}.
\newblock \bibinfo{journal}{\emph{arXiv preprint arXiv:1412.6980}}
  (\bibinfo{year}{2014}).
\newblock


\bibitem[\protect\citeauthoryear{Kit, Webster, Sin, Pan, and Li}{Kit
  et~al\mbox{.}}{2004}]%
        {kit2004clause}
\bibfield{author}{\bibinfo{person}{Chunyu Kit}, \bibinfo{person}{Jonathan~J
  Webster}, \bibinfo{person}{King-Kui Sin}, \bibinfo{person}{Haihua Pan}, {and}
  \bibinfo{person}{Heng Li}.} \bibinfo{year}{2004}\natexlab{}.
\newblock \showarticletitle{Clause alignment for Hong Kong legal texts: A
  lexical-based approach}.
\newblock \bibinfo{journal}{\emph{International Journal of Corpus Linguistics}}
  (\bibinfo{year}{2004}).
\newblock


\bibitem[\protect\citeauthoryear{Koehn, Hoang, Birch, Callison-Burch, Federico,
  Bertoldi, Cowan, Shen, Moran, Zens, et~al\mbox{.}}{Koehn
  et~al\mbox{.}}{2007}]%
        {koehn2007moses}
\bibfield{author}{\bibinfo{person}{Philipp Koehn}, \bibinfo{person}{Hieu
  Hoang}, \bibinfo{person}{Alexandra Birch}, \bibinfo{person}{Chris
  Callison-Burch}, \bibinfo{person}{Marcello Federico}, \bibinfo{person}{Nicola
  Bertoldi}, \bibinfo{person}{Brooke Cowan}, \bibinfo{person}{Wade Shen},
  \bibinfo{person}{Christine Moran}, \bibinfo{person}{Richard Zens},
  {et~al\mbox{.}}} \bibinfo{year}{2007}\natexlab{}.
\newblock \showarticletitle{Moses: Open source toolkit for statistical machine
  translation}. In \bibinfo{booktitle}{\emph{ACL on interactive poster and
  demonstration sessions}}.
\newblock


\bibitem[\protect\citeauthoryear{Li, Xiong, Tu, Zhu, Zhang, and Zhou}{Li
  et~al\mbox{.}}{2017}]%
        {li2017modeling}
\bibfield{author}{\bibinfo{person}{Junhui Li}, \bibinfo{person}{Deyi Xiong},
  \bibinfo{person}{Zhaopeng Tu}, \bibinfo{person}{Muhua Zhu},
  \bibinfo{person}{Min Zhang}, {and} \bibinfo{person}{Guodong Zhou}.}
  \bibinfo{year}{2017}\natexlab{}.
\newblock \showarticletitle{Modeling source syntax for neural machine
  translation}.
\newblock \bibinfo{journal}{\emph{arXiv preprint arXiv:1705.01020}}
  (\bibinfo{year}{2017}).
\newblock


\bibitem[\protect\citeauthoryear{Lin and Wang}{Lin and Wang}{2007}]%
        {lin2007chinese}
\bibfield{author}{\bibinfo{person}{Zhun Lin} {and} \bibinfo{person}{Xiaojie
  Wang}.} \bibinfo{year}{2007}\natexlab{}.
\newblock \showarticletitle{Chinese Ancient-Modern Sentence Alignment}. In
  \bibinfo{booktitle}{\emph{International Conference on Computational
  Science}}.
\newblock


\bibitem[\protect\citeauthoryear{Liu and Wang}{Liu and Wang}{2012}]%
        {liu2012sentence}
\bibfield{author}{\bibinfo{person}{Ying Liu} {and} \bibinfo{person}{Nan Wang}.}
  \bibinfo{year}{2012}\natexlab{}.
\newblock \showarticletitle{Sentence Alignment for Ancient and Modern Chinese
  Parallel Corpus}.
\newblock In \bibinfo{booktitle}{\emph{Emerging Research in Artificial
  Intelligence and Computational Intelligence}}.
\newblock


\bibitem[\protect\citeauthoryear{Luong, Pham, and Manning}{Luong
  et~al\mbox{.}}{2015}]%
        {luong2015effective}
\bibfield{author}{\bibinfo{person}{Minh-Thang Luong}, \bibinfo{person}{Hieu
  Pham}, {and} \bibinfo{person}{Christopher~D Manning}.}
  \bibinfo{year}{2015}\natexlab{}.
\newblock \showarticletitle{Effective approaches to attention-based neural
  machine translation}.
\newblock \bibinfo{journal}{\emph{arXiv preprint arXiv:1508.04025}}
  (\bibinfo{year}{2015}).
\newblock


\bibitem[\protect\citeauthoryear{Mnih and Kavukcuoglu}{Mnih and
  Kavukcuoglu}{2013}]%
        {mnih2013learning}
\bibfield{author}{\bibinfo{person}{Andriy Mnih} {and} \bibinfo{person}{Koray
  Kavukcuoglu}.} \bibinfo{year}{2013}\natexlab{}.
\newblock \showarticletitle{Learning word embeddings efficiently with
  noise-contrastive estimation}. In \bibinfo{booktitle}{\emph{NIPS}}.
\newblock


\bibitem[\protect\citeauthoryear{Nasution, Murakami, and Ishida}{Nasution
  et~al\mbox{.}}{2018}]%
        {nasution2018generalized}
\bibfield{author}{\bibinfo{person}{Arbi~Haza Nasution}, \bibinfo{person}{Yohei
  Murakami}, {and} \bibinfo{person}{Toru Ishida}.}
  \bibinfo{year}{2018}\natexlab{}.
\newblock \showarticletitle{A generalized constraint approach to bilingual
  dictionary induction for low-resource language families}.
\newblock \bibinfo{journal}{\emph{ACM Transactions on Asian and Low-Resource
  Language Information Processing (TALLIP)}} (\bibinfo{year}{2018}).
\newblock


\bibitem[\protect\citeauthoryear{Papineni, Roukos, Ward, and Zhu}{Papineni
  et~al\mbox{.}}{2002}]%
        {Papineni2002BLEU}
\bibfield{author}{\bibinfo{person}{Kishore Papineni}, \bibinfo{person}{Salim
  Roukos}, \bibinfo{person}{Todd Ward}, {and} \bibinfo{person}{Wei-Jing Zhu}.}
  \bibinfo{year}{2002}\natexlab{}.
\newblock \showarticletitle{BLEU: a method for automatic evaluation of machine
  translation}. In \bibinfo{booktitle}{\emph{ACL}}.
\newblock


\bibitem[\protect\citeauthoryear{Schuster and Paliwal}{Schuster and
  Paliwal}{1997}]%
        {schuster1997bidirectional}
\bibfield{author}{\bibinfo{person}{Mike Schuster} {and}
  \bibinfo{person}{Kuldip~K Paliwal}.} \bibinfo{year}{1997}\natexlab{}.
\newblock \showarticletitle{Bidirectional recurrent neural networks}.
\newblock \bibinfo{journal}{\emph{IEEE Transactions on Signal Processing}}
  (\bibinfo{year}{1997}).
\newblock


\bibitem[\protect\citeauthoryear{Sutskever, Vinyals, and Le}{Sutskever
  et~al\mbox{.}}{2014}]%
        {NIPS2014_5346}
\bibfield{author}{\bibinfo{person}{Ilya Sutskever}, \bibinfo{person}{Oriol
  Vinyals}, {and} \bibinfo{person}{Quoc~V Le}.}
  \bibinfo{year}{2014}\natexlab{}.
\newblock \showarticletitle{Sequence to Sequence Learning with Neural
  Networks}.
\newblock In \bibinfo{booktitle}{\emph{NIPS}}.
\newblock


\bibitem[\protect\citeauthoryear{Vaswani, Shazeer, Parmar, Uszkoreit, Jones,
  Gomez, Kaiser, and Polosukhin}{Vaswani et~al\mbox{.}}{2017}]%
        {vaswani2017attention}
\bibfield{author}{\bibinfo{person}{Ashish Vaswani}, \bibinfo{person}{Noam
  Shazeer}, \bibinfo{person}{Niki Parmar}, \bibinfo{person}{Jakob Uszkoreit},
  \bibinfo{person}{Llion Jones}, \bibinfo{person}{Aidan~N Gomez},
  \bibinfo{person}{{\L}ukasz Kaiser}, {and} \bibinfo{person}{Illia
  Polosukhin}.} \bibinfo{year}{2017}\natexlab{}.
\newblock \showarticletitle{Attention is all you need}. In
  \bibinfo{booktitle}{\emph{NIPS}}.
\newblock


\bibitem[\protect\citeauthoryear{Wang and Ren}{Wang and Ren}{2005}]%
        {wang2005chinese}
\bibfield{author}{\bibinfo{person}{Xiaojie Wang} {and} \bibinfo{person}{Fuji
  Ren}.} \bibinfo{year}{2005}\natexlab{}.
\newblock \showarticletitle{Chinese-japanese clause alignment}. In
  \bibinfo{booktitle}{\emph{International Conference on Intelligent Text
  Processing and Computational Linguistics}}.
\newblock


\bibitem[\protect\citeauthoryear{Wu, Schuster, Chen, Le, Norouzi, Macherey,
  Krikun, Cao, Gao, Macherey, et~al\mbox{.}}{Wu et~al\mbox{.}}{2016}]%
        {wu2016google}
\bibfield{author}{\bibinfo{person}{Yonghui Wu}, \bibinfo{person}{Mike
  Schuster}, \bibinfo{person}{Zhifeng Chen}, \bibinfo{person}{Quoc~V Le},
  \bibinfo{person}{Mohammad Norouzi}, \bibinfo{person}{Wolfgang Macherey},
  \bibinfo{person}{Maxim Krikun}, \bibinfo{person}{Yuan Cao},
  \bibinfo{person}{Qin Gao}, \bibinfo{person}{Klaus Macherey}, {et~al\mbox{.}}}
  \bibinfo{year}{2016}\natexlab{}.
\newblock \showarticletitle{Google's neural machine translation system:
  Bridging the gap between human and machine translation}.
\newblock \bibinfo{journal}{\emph{arXiv preprint arXiv:1609.08144}}
  (\bibinfo{year}{2016}).
\newblock


\bibitem[\protect\citeauthoryear{Yang, Zhaopeng, Meng, Cheng, and Zhai}{Yang
  et~al\mbox{.}}{2018}]%
        {liu2018towards}
\bibfield{author}{\bibinfo{person}{Liu Yang}, \bibinfo{person}{Tu Zhaopeng},
  \bibinfo{person}{Fandong Meng}, \bibinfo{person}{Yong Cheng}, {and}
  \bibinfo{person}{Junjie Zhai}.} \bibinfo{year}{2018}\natexlab{}.
\newblock \showarticletitle{Towards Robust Neural Machine Translation}. In
  \bibinfo{booktitle}{\emph{ACL}}.
\newblock


\bibitem[\protect\citeauthoryear{Zhang, Li, and Sun}{Zhang
  et~al\mbox{.}}{2018}]%
        {zhang2018automatic}
\bibfield{author}{\bibinfo{person}{Zhiyuan Zhang}, \bibinfo{person}{Wei Li},
  {and} \bibinfo{person}{Xu Sun}.} \bibinfo{year}{2018}\natexlab{}.
\newblock \showarticletitle{Automatic Transferring between Ancient Chinese and
  Contemporary Chinese}.
\newblock \bibinfo{journal}{\emph{arXiv preprint arXiv:1803.01557}}
  (\bibinfo{year}{2018}).
\newblock


\end{thebibliography}

\end{document}